\documentclass[10pt,twocolumn,letterpaper]{article}

\usepackage{cvpr}
\usepackage{times}
\usepackage{epsfig}
\usepackage{graphicx}
\usepackage{amsmath}
\usepackage{amssymb}

\usepackage{booktabs}

\usepackage[breaklinks=true,bookmarks=false]{hyperref}

\cvprfinalcopy 

\def\cvprPaperID{275} 

\begin{document}

\title{Learning From Noisy Large-Scale Datasets With Minimal Supervision}
\author{Andreas Veit$^{1,\thanks{Work done during internship at Google Research.}}$
 \quad Neil Alldrin$^{2}$
 \quad Gal Chechik$^{2}$
 \quad Ivan Krasin$^{2}$
 \quad Abhinav Gupta$^{2,3}$
 \quad Serge Belongie$^{1}$\\
	$^1$ Department of Computer Science \& Cornell Tech, Cornell University\\
        $^2$ Google Inc, 
		$^3$ The Robotics Institute, Carnegie Mellon University
	}

\maketitle

\begin{abstract}
We present an approach to effectively use millions of images with noisy annotations in conjunction with a small subset of cleanly-annotated images to learn powerful image representations. One common approach to combine clean and noisy data is to first pre-train a network using the large noisy dataset and then fine-tune with the clean dataset. We show this approach does not fully leverage the information contained in the clean set. Thus, we demonstrate how to use the clean annotations to reduce the noise in the large dataset before fine-tuning the network using both the clean set and the full set with reduced noise. The approach comprises a multi-task network that jointly learns to clean noisy annotations and to accurately classify images. We evaluate our approach on the recently released Open Images dataset, containing $\sim$9 million images, multiple annotations per image and over 6000 unique classes. For the small clean set of annotations we use a quarter of the validation set with $\sim$40k images. Our results demonstrate that the proposed approach clearly outperforms direct fine-tuning across all major categories of classes in the Open Image dataset. Further, our approach is particularly effective for a large number of classes with wide range of noise in annotations (20-80\% false positive annotations).

\end{abstract}

\section{Introduction}
Deep convolutional neural networks (ConvNets) proliferate in current machine vision. One of the biggest bottlenecks in scaling their learning is the need for massive and clean collections of semantic annotations for images.
Today, even after five years of success of ImageNet~\cite{Deng2009ImageNetAL}, there is still no publicly available dataset containing an order of magnitude more clean labeled data. To tackle this bottleneck, other training paradigms have been explored aiming to bypass the need of training with expensive manually collected annotations. Examples include unsupervised learning~\cite{Le2012BuildingHF}, self-supervised learning~\cite{Doersch2015,Pinto2016,Zhang2016,Wang2015} and learning from noisy annotations~\cite{Chen2015, Natarajan2013LearningWN}. 
\begin{figure}[t]
\includegraphics[width=1\linewidth]{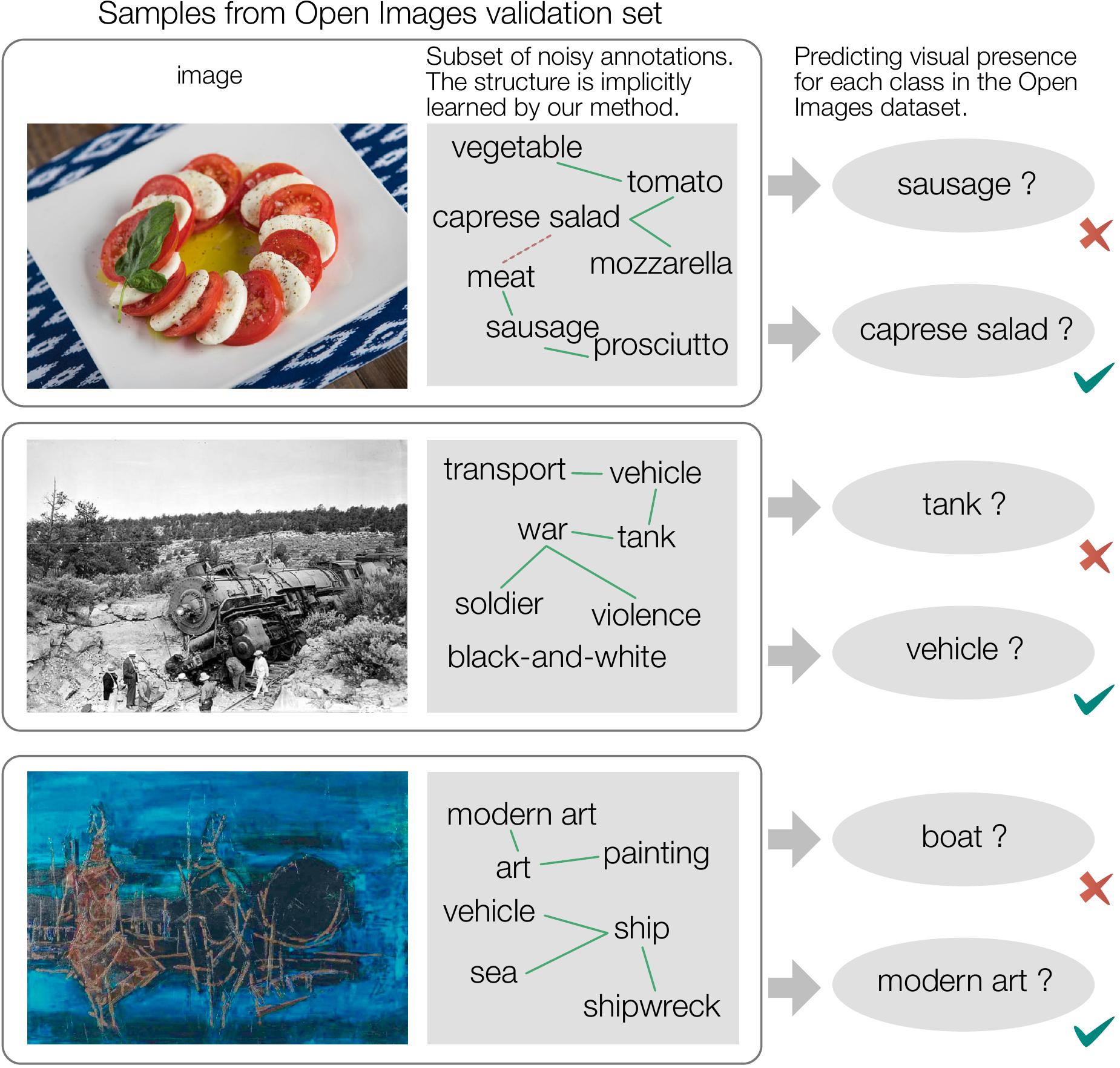}
\caption{Sample images and annotations from the Open Images validation set illustrating the variety of images and the noise in the annotations. We are concerned with the task of training a robust multi-label image classifier from the noisy annotations. While the image annotations are simple lists of classes, our model implicitly learns the structure in the label space. For illustrative purposes, the structure is sketched as a graph with green and red edges denoting strong positive and negative relations. Our proposed approach produces both a cleaned version of the dataset as well as a robust image classifier.}
\label{fig:fig1}
\end{figure}

Most of these approaches make a strong assumption that all annotations are noisy, and no clean data is available. In reality, typical learning scenarios are closer to semi-supervised learning: images have noisy or missing annotations, and a small fraction of images also have clean annotations. This is the case for example, when images with noisy annotations are mined from the web, and then a small fraction gets sent to costly human verification.

In this paper, we explore how to effectively and efficiently leverage a small amount of clean annotations in conjunction with large amounts of noisy annotated data, in particular to train convolutional neural networks. 
One common approach is to pre-train a network with the noisy data and then fine-tune it with the clean dataset to obtain better performance. We argue that this approach does not fully leverage the information contained in the clean annotations.
\begin{figure}[t]
\includegraphics[width=1\linewidth]{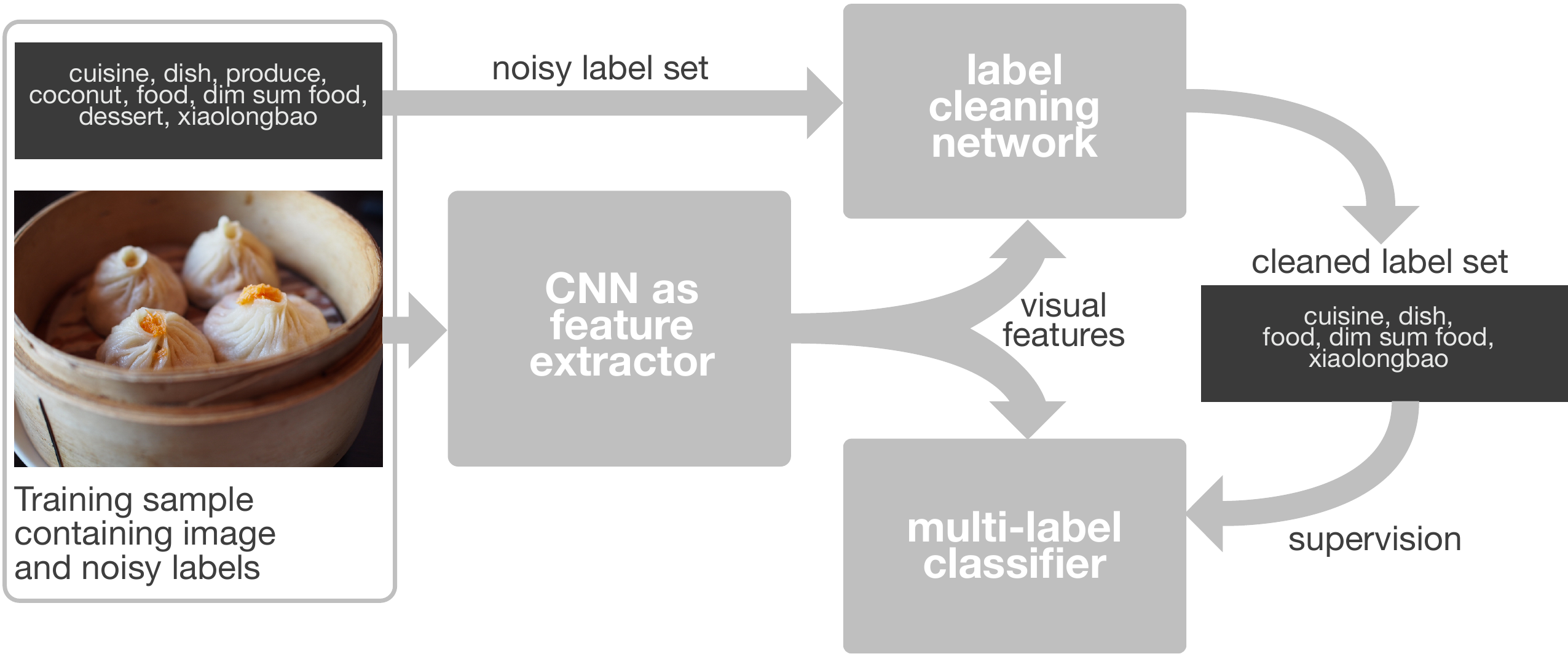}
\caption{High-level overview of our approach. Noisy input labels are cleaned and then used as targets for the final classifier. The label cleaning network and the multi-label classifier are jointly trained and share visual features from a deep convnet. The cleaning network is supervised by the small set of clean annotations (not shown) while the final classifier utilizes both the clean data and the much larger noisy data.}
\label{fig:overview}
\end{figure}
We propose an alternative approach: instead of using the small clean dataset to learn visual representations directly, we use it to learn a mapping between noisy and clean annotations. We argue that this mapping not only learns the patterns of noise, but it also captures the structure in the label space. The learned mapping between noisy and clean annotations allows to clean the noisy dataset and fine-tune the network using both the clean and the full dataset with reduced noise. The proposed approach comprises a multi-task network that jointly learns to clean noisy annotations and to accurately classify images, Figure~\ref{fig:overview}.

In particular, we consider an image classification problem with the goal of annotating images with all concepts present in the image. When considering label noise, two aspects are worth special attention. First, many multi-label classification approaches assume that classes are independent. However, the label space is typically highly structured as illustrated by the examples in Figure~\ref{fig:fig1}. We therefore model the label-cleaning network as conditionally dependent on all noisy input labels. Second, many classes can have multiple \emph{semantic modes}. For example, the class coconut may be assigned to an image containing a drink, a fruit or even a tree. To differentiate between these modes, the input image itself needs to be taken into account. Our model therefore captures the dependence of annotation noise on the input image by having the learned cleaning network conditionally dependent on image features. 

We evaluate the approach on the recently-released large-scale \emph{Open Images Dataset}~\cite{openimages}. The results demonstrate that the proposed approach significantly improves performance over traditional fine-tuning methods. Moreover, we show that direct fine-tuning sometimes hurts performance when only limited rated data is available. In contrast, our method improves performance across the full range of label noise levels, and is most effective for classes having 20\% to 80\% false positive annotations in the training set. The method performs well across a range of categories, showing consistent improvement on classes in all eight high-level categories of Open Images (vehicles, products, art, person, sport, food, animal, plant).

This paper makes the following contributions. First, we introduce a semi-supervised learning framework for multi-label image classification that facilitates small sets of clean annotations in conjunction with massive sets of noisy annotations. Second, we provide a first benchmark on the recently released Open Images Dataset. Third, we demonstrate that the proposed learning approach is more effective in leveraging small labeled data than traditional fine-tuning.

\section{Related Work}
This paper introduces an algorithm to leverage a large corpus of noisily labeled training data in conjunction with a small set of clean labels to train a multi-label image classification model. Therefore, we restrict this discussion to learning from noisy annotations in image classification. For a comprehensive overview of label noise taxonomy and noise robust algorithms we refer to \cite{frenay2014classification}. 

Approaches to learn from noisy labeled data can generally be categorized into two groups: Approaches in the first group aim to directly learn from noisy labels and focus mainly on noise-robust algorithms, e.g.,~\cite{Beigman2009LearningWA,joulin2016learning,Manwani2013NoiseTU}, and label cleansing methods to remove or correct mislabeled data, e.g.,~\cite{Brodley1999IdentifyingMT}. Frequently, these methods face the challenge of distinguishing difficult from mislabeled training samples.
Second, semi-supervised learning (SSL) approaches tackle these shortcomings by combining the noisy labels with a small set of clean labels~\cite{zhu2005semi}. SSL approaches use label propagration such as constrained bootstrapping~\cite{Chen2013} or graph-based approaches~\cite{fergus2009semi}.
Our work follows the semi-supervised paradigm, however focusing on learning a mapping between noisy and clean labels and then exploiting the mapping for training deep neural networks.
\begin{figure*}[t]
\centering
\includegraphics[width=1\linewidth]{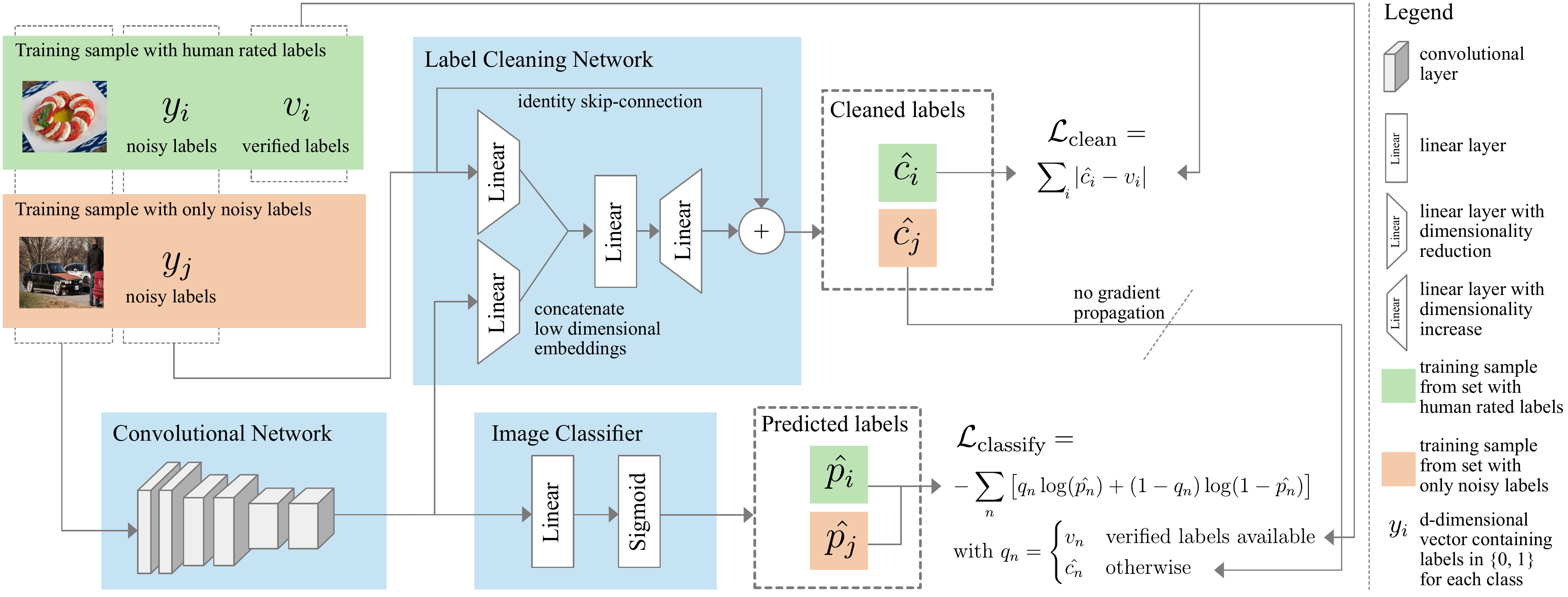} 
\caption{
Overview of our approach to train an image classifier from a very large set of training samples with noisy labels (orange) and a small set of samples which additionally have human verification (green). The model contains a label cleaning network that learns to map noisy labels to clean labels, conditioned on visual features from an Inception V3 ConvNet. The label cleaning network is supervised by the human verified labels and follows a residual architecture so that it only needs to learn the difference between the noisy and clean labels. The image classifier shares the same visual features and learns to directly predict clean labels supervised by either (a) the output of the label cleaning network or (b) the human rated labels, if available.}
\label{fig:architecture}
\end{figure*}

Within the field of training deep neural networks there are three streams of research related to our work. First, various methods have been proposed to explicitly model label noise with neural networks. Natarajan et al.~\cite{Natarajan2013LearningWN} and Sukhbaatar et al.~\cite{sukhbaatar2015training} both model noise that is conditionally independent from the input image. This assumption does not take into account the input image and is thus not able to distinguish effectively between different visual modes and related noise. The closest work in this stream of research is from Xiao et al.~\cite{Xiao2015LearningFM} that proposes an image-conditioned noise model. They first aim to predict the type of noise for each sample (out of a small set of types: no noise, random noise, structured label swapping noise) and then attempt to remove it. Our proposed model is also conditioned on the input image, but differs from these approaches in that it does not explicitly model specific types of noise and is designed for multiple labels per image, not only single labels. Also related is the work of Misra et al.~\cite{MisraNoisy16} who model noise arising from missing, but visually present labels. While their method is conditioned on the input image and is designed for multiple labels per image, it does not take advantage of cleaned labels and their focus is on missing labels, while our approach can address both incorrect and missing labels.

Second, transfer learning has become common practice in modern computer vision. There, a network is pre-trained on a large dataset of labeled images, say ImageNet, and then used for a different but related task, by fine-tuning on a small dataset for specific tasks such as image classification and retrieval~\cite{sharif2014cnn} and image captioning~\cite{Vinyals2015ShowAT}. Unlike these works, our approach aims to train a network from scratch using noisy labels and then facilitates a small set of clean labels to fine-tune the network. 

Third, the proposed approach has surface resemblance to student-teacher models and model compression, where a student, or compressed, model learns to imitate a teacher model of generally higher capacity or with privileged information~\cite{Ba2014DoDN,buciluǎ2006model,Hinton2015DistillingTK,LopezPaz2015UnifyingDA}. In our framework, we train a ConvNet with two classifiers on top, a cleaning network and an image classifier, where the output of the cleaning network is the target of the image classifier. The cleaning network has access to the noisy labels in addition to the visual features, which could be considered privileged information. In our setup the two networks are trained in one joint model. 

\section{Our Approach}
Our goal is to train a multi-label image classifier using a large dataset with relatively noisy labels, where additionally a small subset of the dataset has human verified labels available. This setting naturally occurs when collecting images from the web where only a small subset can be verified by experts. Formally, we have a very large training dataset $T$ comprising tuples of noisy labels $y$ and images $\mathcal{I}$, $T=\{(y_i,\mathcal{I}_i), ...\}$, and a small dataset $V$ of triplets of verified labels $v$, noisy labels $y$ and images $\mathcal{I}$, $V=\{(v_i, y_i, \mathcal{I}_i), ...\}$. The two sets differ significantly in size with $|T| \gg |V|$. For instance, in our experiments, $T$ exceeds $V$ by three orders of magnitude. Each label $y$ or $v$ is a sparse $d$-dimensional vector with a binary annotation for each of the $d$ classes indicating whether it is present in the image or not. Since the labels in $T$ contain significant label noise and $V$ is too small to train a ConvNet, our goal is to design an efficient and effective approach to leverage the quality of the labels in $V$ and the size of $T$. 

\subsection{Multi-Task Label Cleaning Architecture}
We propose a multi-task neural network architecture that jointly learns to reduce the label noise in $T$ and to annotate images with accurate labels. An overview of the model architecture is given in Figure~\ref{fig:architecture}. The model comprises a fully convolutional neural network~\cite{fukushima1980neocognitron,lecun1998gradient,long2015fully} $f$ with two classifiers $g$ and $h$. The first classifier is a \emph{label cleaning network} denoted as $g$ that models the structure in the label space and learns a mapping from the noisy labels $y$ to the human verified labels $v$, conditional on the input image. We denote the \emph{cleaned labels} output by $g$ as $\hat{c}$ so that $\hat{c} = g\left(y,\mathcal{I}\right)$.
The second classifier is an \emph{image classifier} denoted as $h$ that learns to annotate images by imitating the first classifier $g$ by using $g$'s predictions as ground truth targets. We denote the \emph{predicted labels} output by $h$ as $\hat{p}$ so that $\hat{p} = h\left(\mathcal{I}\right)$.

The image classifier $h$ is shown in the bottom row of Figure~\ref{fig:architecture}. First, a sample image is processed by the convolutional network to compute high level image features. Then, these features are passed through a fully-connected layer $w$ followed by a sigmoid $\sigma$, $h = \sigma(w(f(\mathcal{I})))$. The image classifier outputs $\hat{p}$, a $d$-dimensional vector $[0,1]^d$ encoding the likelihood of the visual presence of each of the $d$ classes. 

The label cleaning network $g$ is shown in the top row of Figure~\ref{fig:architecture}. In order to model the label structure and noise conditional on the image, the network has two separate inputs, the noisy labels $y$ as well as the visual features $f(\mathcal{I})$.
The sparse noisy label vector is treated as a bag of words and projected into a low dimensional label embedding that encodes the set of labels. The visual features are similarly projected into a low dimensional embedding. To combine the two modalities, the embedding vectors are concatenated and transformed with a hidden linear layer followed by a projection back into the high dimensional label space. 

Another key detail of the label cleaning network is an identity-skip connection that adds the noisy labels from the training set to the output of the cleaning module. The skip connection is inspired by the approach from He et al.~\cite{he2015deep} but differs in that the residual cleaning module has the visual features as side input. Due to the residual connection, the network only needs to learn the difference between the noisy and clean labels instead of regressing the entire label vector. This simplifies the optimization and enables the network to predict reasonable outputs right from the beginning. When no human rated data is available, the label cleaning network defaults to not changing the noisy labels. As more verified groundtruth becomes available, the network gracefully adapts and cleans the labels. To remain in the valid label space the outputs are clipped to 0 and 1. Denoting the residual cleaning module as $g'$, the label cleaning network $g$ computes cleaned labels
\begin{equation}
\hat{c} = \texttt{clip}(y + g'(y,f(\mathcal{I})),[0,1])
\end{equation}

\subsection{Model Training}
\label{sec:modeltraining}
To train the proposed model we formulate two losses that we minimize jointly using stochastic gradient descent: a label cleaning loss $\mathcal{L}_{\text{clean}}$ that captures the quality of the cleaned labels $\hat{c}$ and a classification loss $\mathcal{L}_{\text{classify}}$ that captures the quality of the predicted labels $\hat{p}$. The calculation of the loss terms is illustrated on the right side of Figure~\ref{fig:architecture}.

The label cleaning network is supervised by the verified labels of all samples $i$ in the human rated set $V$. The cleaning loss is based on the difference between the cleaned labels $\hat{c_i}$ and the corresponding ground truth verified labels~$v_i$, 
\begin{equation}
\mathcal{L}_{\text{clean}} = \sum_{i \in V}|\hat{c_i}-v_i|
\end{equation}
We choose the absolute distance as error measure, since the label vectors are very sparse. Other measures such as the squared error tend to smooth the labels. 

For the image classifier, the supervision depends on the source of the training sample. For all samples $j$ from the noisy dataset $T$, the classifier is supervised by the cleaned labels $\hat{c_j}$ produced by the label cleaning network. For samples $i$ where human ratings are available, $i \in V$, supervision comes directly from the verified labels $v_i$. To allow for multiple annotations per image, we choose the cross-entropy as classification loss to capture the difference between the predicted labels $\hat{p}$ and the target labels.
\begin{equation}
\begin{split}
\mathcal{L}_{\text{classify}} = 
&-\sum_{j \in T}\big[\hat{c_j}\log(\hat{p_j}) + (1-\hat{c_j})\log(1-\hat{p_j}) \big] \\ 
&-\sum_{i \in V}\big[v_i\log(\hat{p_i}) + (1-v_i)\log(1-\hat{p_i}) \big] 
\end{split}
\label{equ:classifyloss}
\end{equation}
\begin{table}[t]
\centering
\caption{\label{tab:verticals}Breakdown of the ground-truth annotations in the validation set of the Open Images Dataset by high-level category. The dataset spans a wide range of everyday categories from manmade products to personal activities as well as coarse and fine-grained natural species.}
\begin{tabular}{@{}lll@{}} \toprule
    \text{high-level category} & \text{unique labels}& \text{annotations} \\ \midrule
    \text{vehicles}  & 944 & 240,449\\ 
    \text{products}  & 850 & 132,705\\
    \text{art}  & 103 & 41,986\\
    \text{person}  & 409 & 55,417\\
    \text{sport}  & 446 & 65,793\\
	\text{food}  & 862 & 140,383\\
    \text{animal}  & 1064 & 187,147\\
    \text{plant}  & 517 & 87,542 \\ 
    \text{others}  & 1388 & 322,602 \\\bottomrule
\end{tabular}
\end{table}

It is worth noting that the vast majority of training examples come from set $T$. Thus, the second summation in Equation~\ref{equ:classifyloss} dominates the overall loss of the model. To prevent a trivial solution, in which the cleaning network and classifier both learn to predict label vectors of all zeros, $\hat{c_j}=\hat{p_j}=\{0\}^d$, the classification loss is only propagated to $\hat{p_j}$. The cleaned labels $\hat{c_j}$ are treated as constants with respect to the classification and only incur gradients from the cleaning loss.

To train the cleaning network and image classifier jointly we sample training batches that contain samples from $T$ as well as $V$ in a ratio of $9:1$. This allows us to utilize the large number of samples in $T$ while giving enough supervision to the cleaning network from $V$.

\section{Experiments}
\subsection{Dataset}
We evaluate our proposed model on the recently-released Open Images dataset~\cite{openimages}. The dataset is uniquely suited for our task as it contains a very large collection of images with relatively noisy annotations and a small validation set with human verifications. The dataset is multi-label and massively multi-class in the sense that each image contains multiple annotations and the vocabulary contains several thousand unique classes. In particular, the training set contains 9,011,219 images with a total of 79,156,606 annotations, an average of 8.78 annotations per image. The validation set contains another 167,056 images with a total of 2,047,758 annotations, an average of 12.26 annotations per image.  The dataset contains 6012 unique classes and each class has at least 70 annotations over the whole dataset. 
\begin{figure}[t]
\begin{center}
\begin{tabular}{@{}c@{}c@{}}
\includegraphics[width=0.5\linewidth]{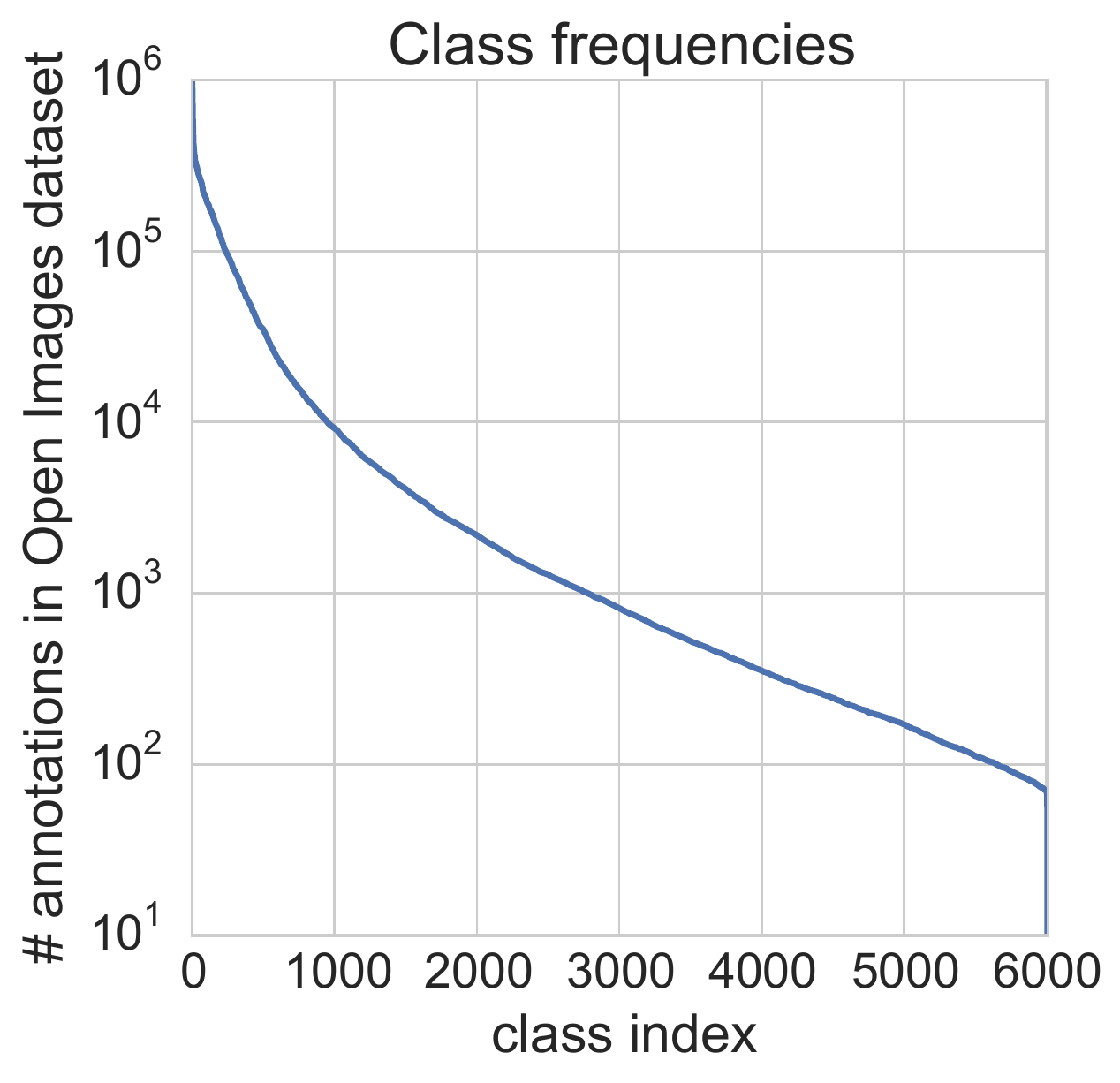} &
\includegraphics[width=0.5\linewidth]{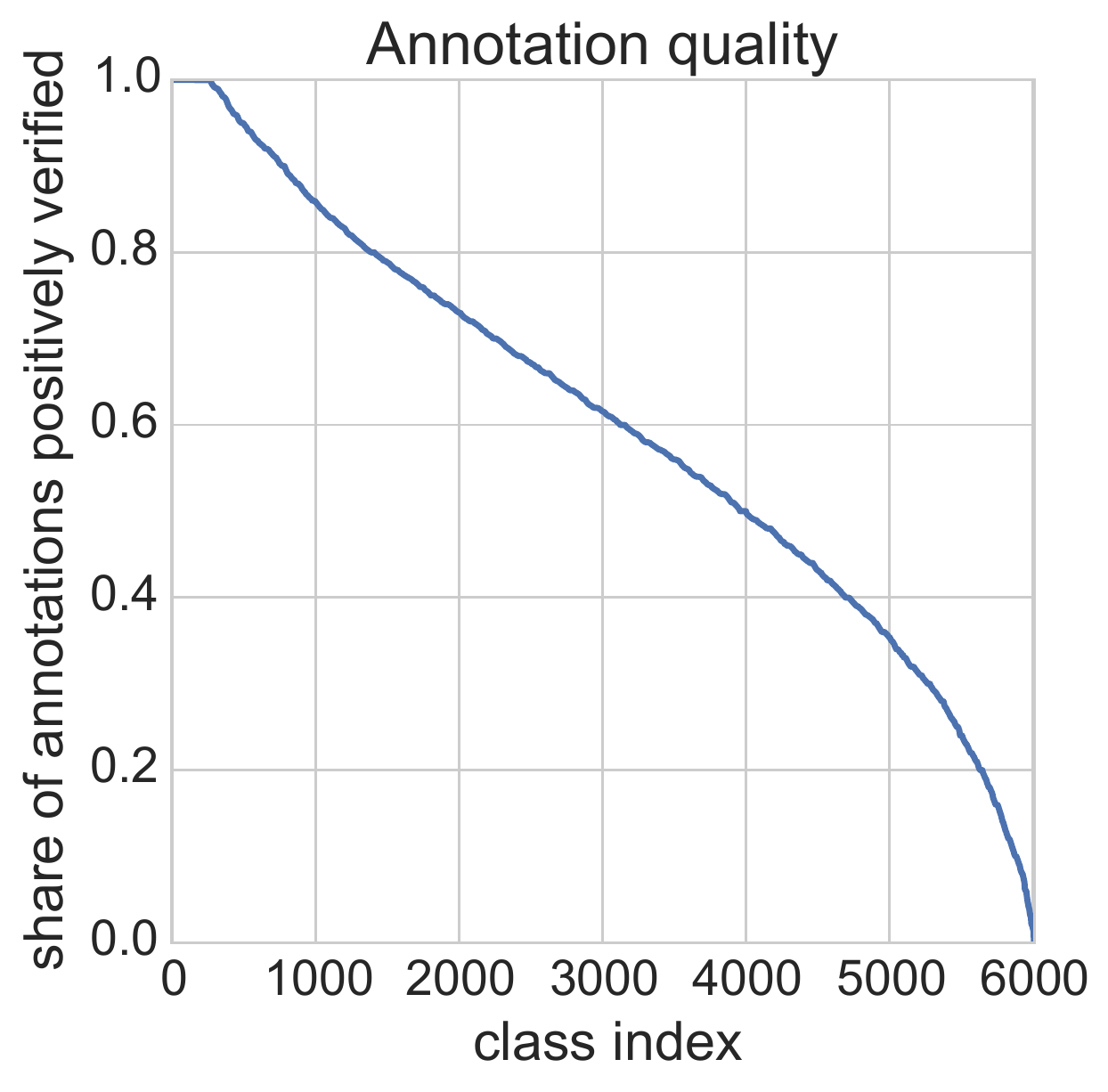} \\
(a) Class frequencies &
(b) Annotation quality
\end{tabular}
\end{center}
\vspace{-5pt}
\caption{\label{fig:stats}Label statistics for the Open Images dataset. Classes are ordered by frequency and annotation quality respectively. (a) Classes are heavily skewed in terms of number of annotations, e.g., "vehicle" occurs over 900,000 times whereas "honda nsx" only occurs 70 times. (b) Classes also vary significantly in annotation quality which refers to the probability that an image labeled with a class actually contains that class. Overall, more than 70\% of the $\sim$80M annotations in the dataset are correct and common classes tend to have higher annotation quality.}
\vspace{-10pt}
\end{figure}

One key distinction from other datasets is that the classes in Open Images are not evenly distributed. Some high-level classes such as `vehicle` have over 900,000 annotations while many fine-grain classes are very sparse, e.g.,~`honda nsx` only occurs 70 times. Figure~\ref{fig:stats}(a) shows the distribution of class frequencies over the validation set. 
Further, many classes are highly related to each other. To differentiate our evaluation between clusters of semantically closely related classes, we group classes with respect to their associated high-level category. Table~\ref{tab:verticals} gives an overview of the main categories and their statistics over the validation set. 

Besides the uneven distribution of classes, another key distinction of the dataset is annotation noise. The training ground-truth comes from an image classifier similar to Google Cloud Vision API\footnote{\text{https://cloud.google.com/vision/}}. Due to the automated annotation process, the training set contain a considerable amount of noise.  Using the validation set to estimate the annotation quality, we observe that 26.6\% of the automatic annotations are considered false positives. The quality varies widely between the classes. Figure~\ref{fig:stats}(b) shows the distribution of the quality of the automated annotations. While some classes only have correct annotations, others do not have any. However, the noise is not random, since the label space is highly structured, see Figure~\ref{fig:fig1} for examples. 

For our experiments, we use the training set as large corpus of images with only noisy labels $T$. Further, we split the validation set into two parts: one quarter of about 40 thousand images is used in our cleaning approach providing both noisy and human verified labels $V$. The remaining three-quarters are held out and used only for validation.

\subsection{\label{section:metrics}Evaluation Task and Metrics}
We evaluate our approach using multi-label image classification, i.e.,\ predicting a score for each class-image pair indicating the likelihood the concept described by the class is present in the image. 

There is no standard evaluation procedure yet for classification on the Open Images dataset. Thus, we choose the widely used average precision (AP) as metric to evaluate performance. The AP for each class $c$ is
\begin{equation}
AP_c = \frac{\sum_{k = 1}^{N}\text{Precision}(k,c)\cdot \text{rel}(k,c)}{\text{number of positives}}
\end{equation}
where Precision($k,c$) is the precision for class $c$ when retrieving $k$ annotations and rel($k,c$) is an indicator function that is $1$ iff the ground truth for class $c$ and the image at rank $k$ is positive. $N$ is the size of the validation set. We  report the mean average precision (MAP) that takes the average over the APs of all $d$, 6012, classes, $MAP = 1/d \sum_{c=1}^{d}AP_c$. 
Further, because we care more about the model performance on commonly occurring classes we also report a class agnostic average precision, $AP_{all}$. This metric considers every annotation equally by treating them as coming from one single class.

Evaluation on Open Images comes with the challenge that the validation set is collected by verifying the automatically generated annotations. As such, human verification only exists for a subset of the classes for each image. This raises the question of how to treat classes without verification. One option is to consider classes with missing human-verification as negative examples. However, we observe that a large number of the highly ranked annotations are likely correct but not verified. Treating them as negatives would penalize models that differ substantially from the model used to annotate the dataset. Thus, we choose instead to ignore classes without human-verification in our metrics. This means the measured precision at full recall for all approaches is very close to the precision of the annotation model, see the PR curve in Figure~\ref{fig:prcurves}(a).

\begin{table}[t]
\centering
\caption{\label{tab:results_overall}Comparison of models in terms of AP and MAP on the held out subset of the Open Images validation set. Our approach outperforms competing methods. See Sections~\ref{section:metrics} and~\ref{section:baselines} for more details on the metrics and model variants.}
\begin{tabular}{@{}lll@{}} \toprule
    Model & $AP_{all}$ & $MAP$ \\ \midrule
    Baseline & 
83.82  &
61.82\\
Misra et al.~\cite{MisraNoisy16} visual classifier & 
83.55  &
61.85\\
Misra et al.~\cite{MisraNoisy16} relevance classifier & 
83.79  &
61.89\\
Fine-Tuning with mixed labels & 
84.80 &
61.90\\
Fine-Tuning with clean labels & 
85.88 & 
61.53\\
\textbf{Our Approach} with pre-training & 
\textbf{87.68} & 
62.36\\
\textbf{Our Approach} trained jointly & 
87.67 &
\textbf{62.38}\\ \bottomrule
\end{tabular}
\vspace{-7pt}
\end{table}

\subsection{\label{section:baselines}Baselines and Model Variants}
As baseline model for our evaluation we train a network solely on the noisy labels from the training set. We refer to this model as \textbf{baseline} and use it as the starting point for all other variants. We compare the following approaches.

\noindent
\textbf{Fine-tune with clean labels}: A common approach is to use the clean labels directly to supervise the last layer. This approach converges quickly because the dataset for fine-tuning is very small; however, many classes have very few training samples making it prone to overfitting. 
\begin{figure}[t]
\centering
\includegraphics[width=0.9\linewidth]{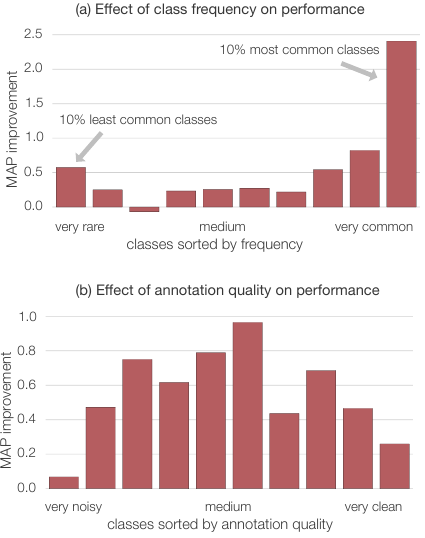}
\caption{Performance gain of our approach with respect to how common a class is and how noisy its annotations are in the dataset. We sort the classes along the x-axis, group them into 10 equally sized groups and compute the MAP gain over the baseline within each group. (a) Most effective is our approach for classes that occur frequently. (b) Our approach improves performance across all levels of annotation quality. It shows the largest gain for classes with 20\% to 80\% false annotations, classes that contain sufficient negative and positive examples in the human rated set.}
\label{fig:qual_impr}
\vspace{-10pt}
\end{figure}

\noindent
\textbf{Fine-tune with mix of clean and noisy labels}: This addresses the shortcomings of limited training samples. We fine-tune the last layer with a mix of training samples from the small clean and the large noisy set (in a 1 to 9 ratio).
\begin{table*}[t]
\centering
\caption{\label{tab:results_verticals_map}Mean average precision for classes grouped according to high-level categories of the Open Images Dataset. Our method consistently performs best across all categories.}
\begin{tabular}{@{}llllllllll@{}} \toprule
    \text{Model} & \text{vehicles} & \text{products} & \text{art} & \text{person} & \text{sport} & \text{food} & \text{animal} & \text{plant}  \\ \midrule
    Baseline & 
56.92 & 
61.51 & 
68.28 & 
59.46 & 
62.84 & 
61.79 & 
61.14 & 
59.00 \\
Fine-Tuning with mixed labels & 
57.00 & 
61.56 & 
68.23 & 
59.49 & 
63.12 & 
61.77 & 
\textbf{61.27} & 
59.14  \\
Fine-Tuning with clean labels & 
56.93 & 
60.94 & 
68.12 & 
58.39 & 
62.56 & 
61.60 & 
61.18 & 
58.90 \\
\textbf{Our Approach} with pre-training & 
57.15 & 
\textbf{62.31} & 
68.89 & 
60.03 & 
63.60 & 
\textbf{61.87} & 
61.26 & 
\textbf{59.45}  \\
\textbf{Our Approach} trained jointly & 
\textbf{57.17} & 
\textbf{62.31} & 
\textbf{68.98} & 
\textbf{60.05} & 
\textbf{63.61} & 
\textbf{61.87} & 
\textbf{61.27} & 
59.36  \\ \bottomrule
\end{tabular}
\end{table*}

\begin{figure*}[t]
\begin{center}
\begin{tabular}{@{}c@{}c@{}c@{}}
\includegraphics[width=0.33\linewidth]{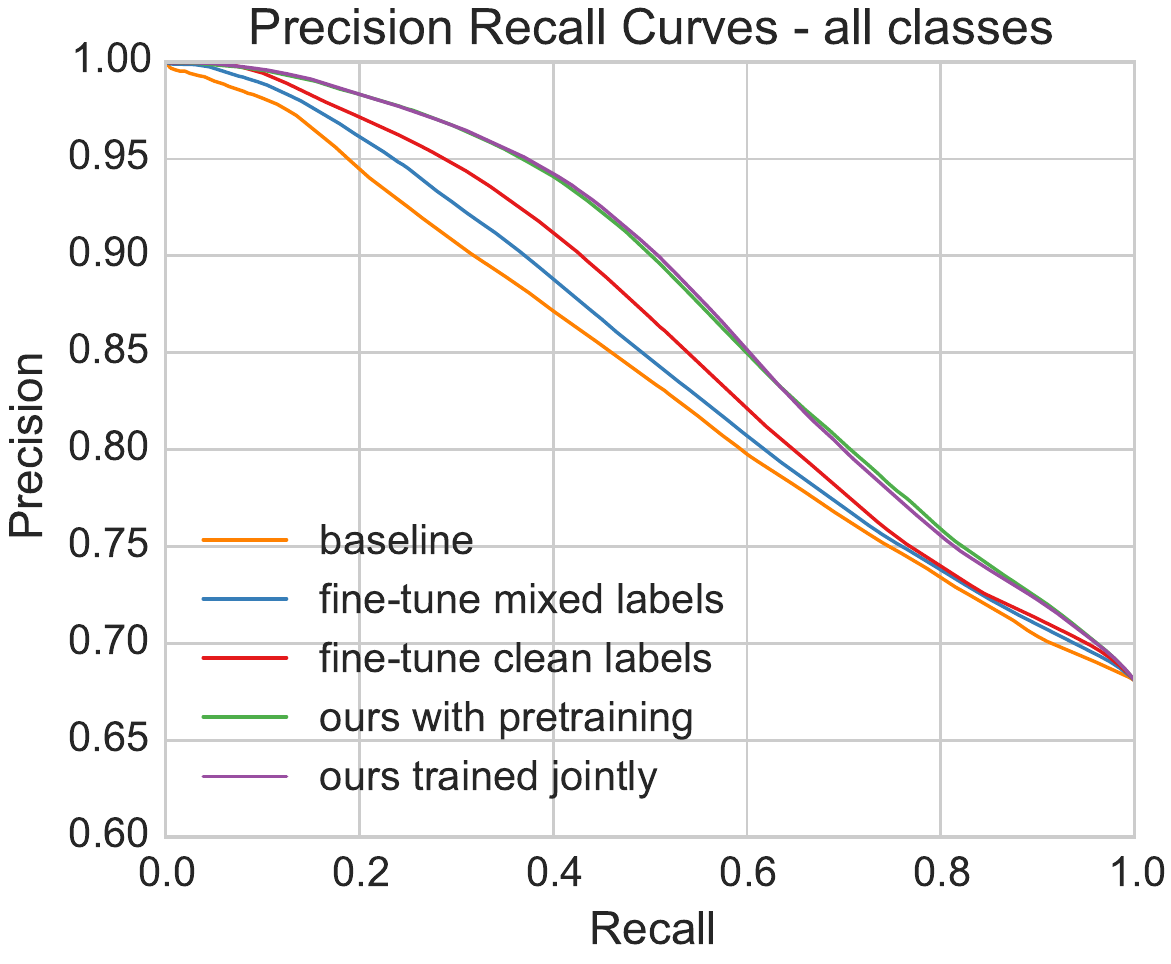} &
\includegraphics[width=0.33\linewidth]{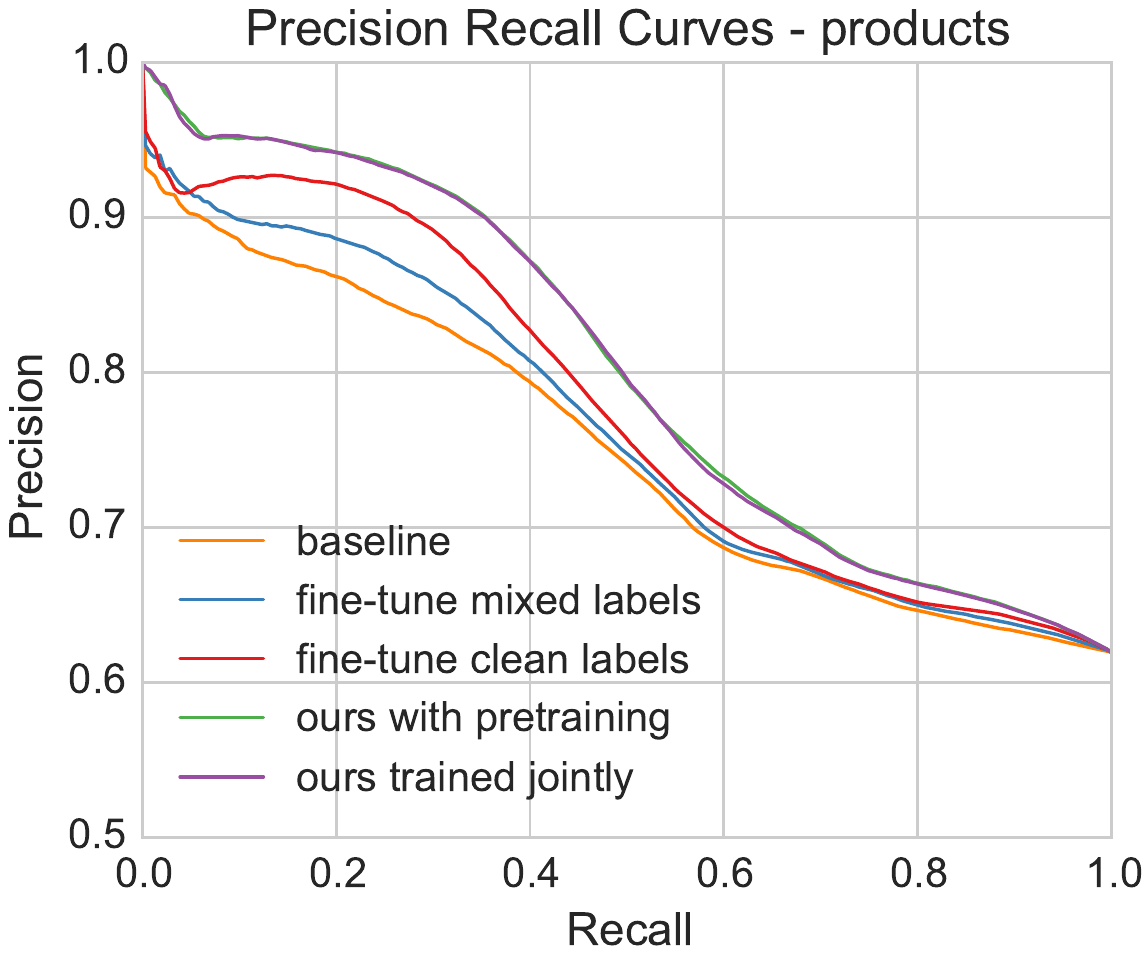} &
\includegraphics[width=0.33\linewidth]{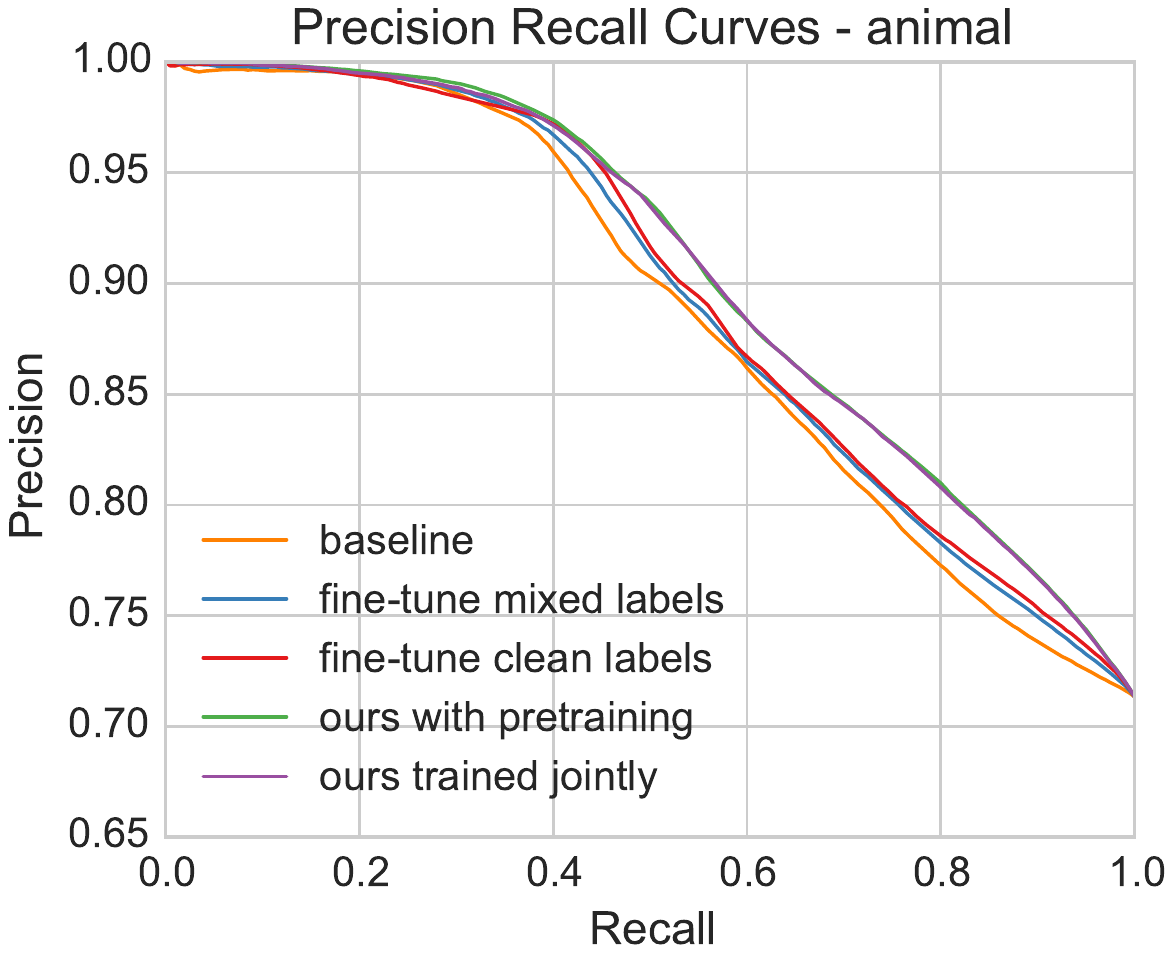} \\
(a) all classes &
(b) products &
(c) animal \\
\end{tabular}
\end{center}
\vspace{-7pt}
\caption{\label{fig:prcurves}Precision-recall curves for all methods measured over all annotations and for the major categories of products and animals. In general, our method performs best, followed by fine-tuning with clean labels, fine-tuning with a mix of clean and noisy labels, and the baseline model. Over all classes, we see improvements across all confidence levels. For products the main improvements come from annotations with high-confidence. For animals we observe mainly gains in the lower confidence regime. It is worthy of note there is virtually no difference between pre-training the cleaning network and learning it jointly.}
\vspace{-10pt}
\end{figure*}

\noindent
\textbf{Our approach with pre-trained cleaning network}: We compare two different variants of our approach. Both are trained as described in Section~\ref{sec:modeltraining}. They only differ with respect to their initialization. For first variant, we initially train just the label cleaning network on the human rated data. Then, subsequently we train the cleaning network and the classification layer jointly.

\noindent
\textbf{Our approach trained jointly}: To reduce the overhead of pre-training the cleaning network, we also train a second variant in which the cleaning network and the classification layer are trained jointly right from the beginning. 

\noindent
\textbf{Misra et al.}: Finally, we compare to the approach of Misra et al.~\cite{MisraNoisy16}. As expected, our method performs better since their model does not utilize the clean labels and their noise model focuses only on missing labels.

\subsection{Training Details}
For our base model, we use an Inception v3 network architecture~\cite{szegedy2015rethinking}, implemented with TensorFlow~\cite{abadi2016tensorflow} and optimized with RMSprop~\cite{tieleman2012lecture} with learning rate $0.045$ and exponential learning rate decay of $0.94$ every $2$ epochs. As only modification to the architecture we replace the final softmax with a 6012-way sigmoid layer. The network is supervised with a binary cross-entropy loss. We trained the baseline model on 50 NVIDIA K40 GPUs using the noisy labels from the Open Images training set. We stopped training after 49 million mini-batches (with 32 images each). This network is the starting point for all model variants.

The four different fine-tuning variants are trained for additional 4 million batches each. The learning rate for the last classification layer is initialized to $0.001$. For the cleaning network it is set higher to $0.015$, because its weights are initialized randomly. For the approach with pre-trained cleaning network, it is first trained with a learning rate of $0.015$ until convergence and then set to $0.001$ once it is trained jointly with the classifier. To balance the losses, we weight $\mathcal{L}_{\text{clean}}$ with $0.1$
 and $\mathcal{L}_{\text{classify}}$ with $1.0$. 

\begin{figure*}[t]
\centering
\includegraphics[width=1\linewidth]{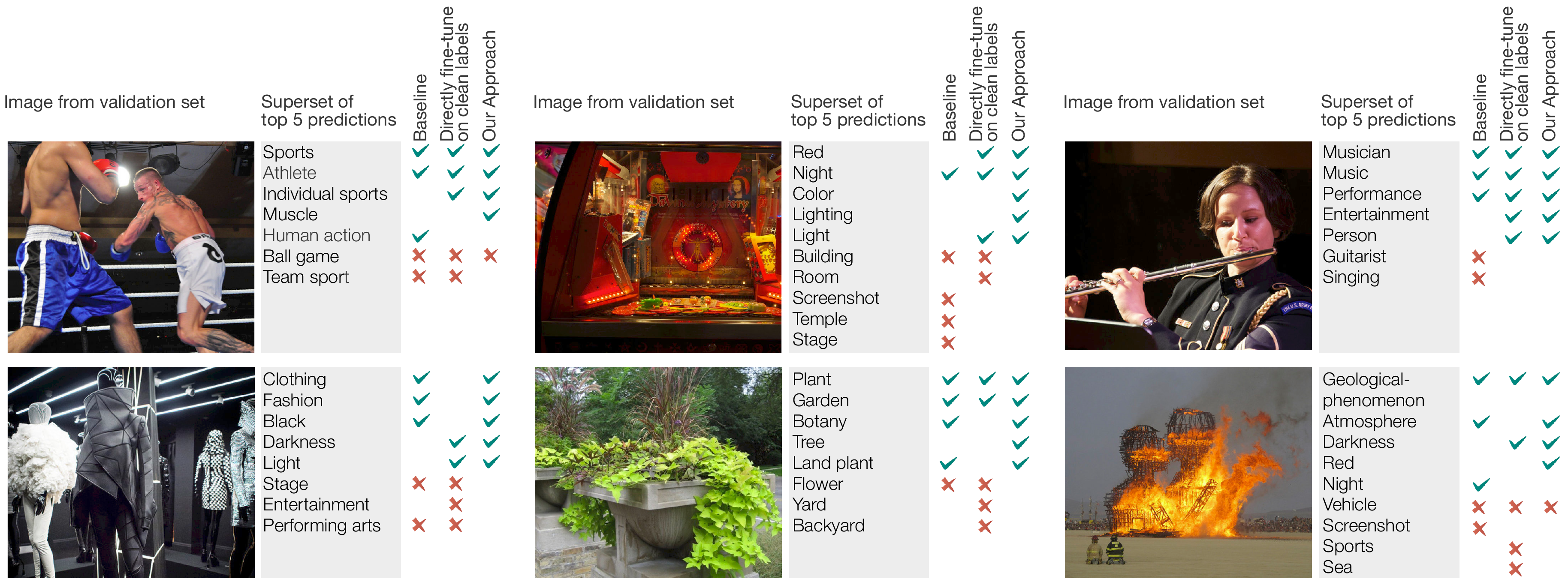}
\caption{Examples from the hold-out portion of the Open Images validation set. We show the top 5 most confident predictions of the baseline model, directly fine-tuning on clean labels and our approach, along with whether the prediction is correct of incorrect. Our approach consistently removes false predictions made by the baseline model. Example gains are the removal of `team sport' and recall of `muscle' in the upper left. This is a very typical example as most sport images are annotated with `ball game' and `team sport' in the dataset. 
Directly fine-tuning achieves mixed results. Sometimes it performs similar to our approach and removes false labels, but for others it even recalls more false labels. This illustrates the challenge of overfitting for directly-finetuning.}
\label{fig:examples}
\vspace{-10pt}
\end{figure*}
\subsection{Results}
We first analyze the overall performance of the proposed approach. Table~\ref{tab:results_overall} shows mean average precision as well as class agnostic average precision. Generally, performance in terms of $AP_{all}$ is higher than for $MAP$, indicating that average precision is higher for common than for rare classes. Considering all annotations equally, $AP_{all}$, we see clear improvements of all variants over the baseline. Further, the two variants of the proposed approach perform very similar and demonstrate a significant lead over direct fine-tuning.

The results in terms of $MAP$ show a different picture. Instead of improving performance, fine-tuning on the clean data directly even hurts the performance. This means the improvement in $AP_{all}$ is due to a few very common classes, but performance in the majority of classes decreases. For many classes the limited number of annotations in the clean label set seems to lead to overfitting. Fine-tuning on clean and noisy annotations alleviates the problem of overfitting, however, at a cost in overall performance. Our approach on the other hand does not face the problem of overfitting. Again, our two variants perform very similar and both demonstrate significant improvements over the baseline and direct fine-tuning. The consistent improvement over all annotations and over all classes shows that our approach is clearly more effective than direct fine-tuning to extract the information from the clean label set.

The similar performance of the variants with and without pre-trained cleaning network indicate that pre-training is not required and our approach can be trained jointly. Figure~\ref{fig:examples} shows example results from the validation set.

\subsubsection{Effect of label frequency and annotation quality}
We take a closer look at how class frequency and annotation quality effects the performance of our approach. 

Figure~\ref{fig:qual_impr}(a) shows the performance improvement of our approach over the baseline with respect to how common a class is. The $x$-axis shows the 6012 unique classes in increasing order from rare to common. We group the classes along the axis into 10 equally sized groups
The result reveals that our approach is able to achieve performance gains across almost all levels of frequency. Our model is most effective for very common classes and shows improvement for all but a small subset of rare classes. Surprisingly, for very rare classes, mostly fine-grained object categories, we again observe an improvement.

Figure~\ref{fig:qual_impr}(b) shows the performance improvement with respect to the annotation quality. The x-axis shows the classes in increasing order from very noisy annotations to always correct annotations. Our approach improves performance across all levels of annotation quality. The largest gains are for classes with medium levels of annotation noise. For classes with very clean annotations the performance is already very high, limiting the potential for further gains. For very noisy classes nearly all automatically generated annotations are incorrect. This means the label cleaning network receives almost no supervision for what a positive sample is. Classes with medium annotation quality contain sufficient negative as well as positive examples in the human rated set and have potential for improvement.

\subsubsection{Performance on high-level categories of Open Images dataset}
Now we evaluate the performance on the major sub-categories of classes in the Open Images dataset. The categories, shown in Table~\ref{tab:verticals}, range from man-made objects such as vehicles to persons and activities to natural categories such as plants. 
Table~\ref{tab:results_verticals_map} shows the mean average precision. Our approach clearly improves over the baseline and direct fine-tuning. Similar results are obtained for class agnostic average precision, where we also show the precision-recall curves for the major categories of products and animals in Figure~\ref{fig:prcurves}. For products the main improvements come from high-confidence labels,
whereas, for animals we observe mainly gains in the lower confidence regime.

\section{Conclusion}
How to effectively leverage a small set of clean labels in the presence of a massive dataset with noisy labels? We show that using the clean labels to directly fine-tune a network trained on the noisy labels does not fully leverage the information contained in the clean label set. We present an alternative approach in which the clean labels are used to reduce the noise in the large dataset before fine-tuning the network using both the clean labels and the full dataset with reduced noise. We evaluate on the recently released Open Images dataset showing that our approach outperforms direct fine-tuning across all major categories of classes.

There are a couple of interesting directions for future work. The cleaning network in our setup combines the label and image modalities with a concatenation and two fully connected layers. Future work could explore higher capacity interactions such as bilinear pooling. Further, in our approach the input and output vocabulary of the cleaning network is the same. Future work could aim to learn a mapping of noisy labels in one domain into clean labels in another domain such as Flickr tags to object categories.

\section*{Acknowledgements}
We would like to thank Ramakrishna Vedantam for insightful feedback as well as the AOL Connected Experiences Laboratory at Cornell Tech. This work was funded in part by a Google Focused Research Award.

{\small
\bibliographystyle{ieee}
\bibliography{egbib}
}

\end{document}